\documentclass[a4paper]{article}
\usepackage{hyperref}
\pdfoutput=1
\usepackage{pdfpages}

\usepackage[utf8]{inputenc} 
\usepackage[T1]{fontenc}    
\usepackage{bm}
\usepackage{hyperref}       
\usepackage{url}            
\usepackage{booktabs}       
\usepackage{amsfonts}       
\usepackage{nicefrac}       
\usepackage{microtype}      
\usepackage{amsthm}
\usepackage{mathtools}
\usepackage{amssymb, amsmath}
\usepackage{graphicx}
\usepackage{multicol}
\usepackage{multirow}
\usepackage{subfigure}
\usepackage{CJKutf8}
\usepackage{xcolor}  
\usepackage[numbers]{natbib}
\newtheorem{definition}{Definition}
\newtheorem{theorem}{Theorem}
\newtheorem{lemma}{Lemma}
\newtheorem{corollary}{Corollary}
\newtheorem{remark}{Remark}

\newcommand{\BlackBox}{\rule{1.5ex}{1.5ex}}

\usepackage[acronym,nomain]{glossaries}

\usepackage[acronym,nomain]{glossaries}
\usepackage{authblk}
\newglossary[slg]{symbolslist}{syi}{syg}{Symbolslist}
\makeglossaries

\makeatletter
\newcommand{\figcaption}[1]{\def\@captype{figure}\caption{#1}}
\newcommand{\tblcaption}[1]{\def\@captype{table}\caption{#1}}
\makeatother

\title{Information-theoretic Analysis of Test Data Sensitivity in Uncertainty}

\author[1]{\href{mailto:Futoshi Futami <futoshi.futami.uk@hco.ntt.co.jp>?Subject=Archiv paper}{Futoshi Futami\thanks{futami.futoshi.es@osaka-u.ac.jp}}{}} 
\author[2]{Tomoharu Iwata}
\affil[1]{%
    Osaka university\\
    Osaka, Japan
}
\affil[2]{%
    Communication Science Laboratories\\
    NTT\\
    Kyoto, Japan
}

\begin{document}

\maketitle

\begin{abstract}
Bayesian inference is often utilized for uncertainty quantification tasks. A recent analysis by \cite{xu2020minimum} rigorously decomposed the predictive uncertainty in Bayesian inference into two uncertainties, called aleatoric and epistemic uncertainties, which represent the inherent randomness in the data-generating process and the variability due to insufficient data, respectively. They analyzed those uncertainties in an information-theoretic way, assuming that the model is well-specified and treating the model's parameters as latent variables. However, the existing information-theoretic analysis of uncertainty cannot explain the widely believed property of uncertainty, known as the sensitivity between the test and training data. It implies that when test data are similar to training data in some sense, the epistemic uncertainty should become small. In this work, we study such uncertainty sensitivity using our novel decomposition method for the predictive uncertainty. Our analysis successfully defines such sensitivity using information-theoretic quantities. Furthermore, we extend the existing analysis of Bayesian meta-learning and show the novel sensitivities among tasks for the first time.
\end{abstract}

\section{Introduction}\label{sec:intro}
Uncertainty qualification for predictions of machine learning algorithms has become increasingly important. Such information is used in the detection of domain shifts \citep{NEURIPS2019_8558cb40}, adversarial attacks \citep{NEURIPS2018_586f9b40}, Bayesian optimization \citep{hernandez2014predictive}, and reinforcement learning \citep{janz2019successor}. 
The Bayesian inference is widely used in such applications since it represents uncertainties through a posterior distribution updated from the prior distribution using training data \citep{PRML}. 
 
 Existing studies \citep{kendall2017uncertainties, bhatt2021uncertainty} often categorized uncertainties into two types; one is aleatoric uncertainty, which is caused by the noise inherent in the data-generating process. The other is called epistemic uncertainty, caused by the lack of data. Recently, \cite{xu2020minimum} have rigorously decomposed the uncertainty of the prediction into aleatoric and epistemic uncertainties by focusing on loss functions in supervised learning. 
  The key idea of their analysis is that assuming the model is well-specified, model parameters are treated as latent variables and marginalized, similarly to Bayesian inference. Thus, they called the setting {\bf Bayesian learning}. Under this setting, the optimal decision rule is obtained on the basis of the Bayesian posterior distribution. 
  
  Under the Bayesian learning setting, they proposed treating the aleatoric uncertainty as the Bayes risk since the noise inherent in the data-generating process is closely related to the fundamental difficulty of learning. As we introduce in Sec.~\ref{sec:Background}, they defined the epistemic uncertainty as the excess risk obtained by subtracting the Bayes risk from the optimal risk under the optimal decision rule since such excess risk corresponds to the ``loss due to lack of data''.  Then, they analyzed the epistemic uncertainty by studying the excess risk. 
 For example, when the risk function is the log loss, the optimal risk becomes the entropy of the posterior predictive distribution, and the Bayes risk corresponds to the entropy of the data-generating distribution. Then, the excess risk is equivalent to {\bf conditional mutual information (CMI)}, see Eq.~\eqref{eq_CMI} for details.
 Thus, the CMI corresponds to epistemic uncertainty. They showed that such CMI monotonically decreases with the training dataset size, a desirable property of epistemic uncertainty. As for other common loss functions, the excess risk can also be upper-bounded by the CMI. These settings have recently been extended to Bayesian meta-learning settings, where we assume a hyperprior distribution on prior distributions \citep{jose2021information}. 

The limitation of these existing Bayesian learning analyses is that they cannot explain the widely believed geometric property of epistemic uncertainty: If the given test data point is similar to the training data in some sense, the uncertainty at such test data point should be small since there is sufficient information for the prediction. On the other hand, if the test data is less similar to the training data, the uncertainty should be large. This property is called the {\bf sensitivity} between the test and training data points. Linear models and Gaussian processes exhibit this property \citep{PRML} since the variance of their posterior predictive distribution explicitly depends on the distance between the test and training data under the given feature map. This sensitivity property has recently received attention in deep learning methods \citep{liu2020simple,van2020uncertainty}. However, existing analysis under Bayesian learning cannot explain such sensitivity for the test data. Since \cite{xu2020minimum} analyzed the CMI by upper-bounding it using the mutual information (MI) between the training data and model parameters, and such MI does not contain the information of the test data point, see Eq.~\eqref{eq_CMI_MI} for details.

In this paper, we continue the uncertainty analysis under Bayesian learning and aim to analyze the sensitivity between the test and training data points. To achieve this, we first present the novel decomposition of the CMI in Theorem~\ref{thm_main_cmd}. Using this, we formally define the sensitivity between the test and training data using an information-theoretic quantity. Then we provide the theoretical and numerical analyses of this quantity. We also apply our analysis to the meta-learning setting similarly to \cite{jose2021information} and determined the sensitivity between the meta-training and meta-test tasks in Theorem~\ref{thm_main_cmd_meta}. To the best of our knowledge, the sensitivity among tasks is presented for the first time.

Another contribution of this work is a new information-theoretic upper bound of the CMI, which includes the interaction between training data points in Corollary~\ref{col_improved_bound}. Our new bound is tighter than the existing bound proposed by \cite{xu2020minimum}. Finally, we present a new exact characterization of the generalization error using our novel decomposition in Theorem~\ref{eq_bayesian_generalization} and show a new connection to the existing information-theoretic generalization error bounds under the frequentist setting \citep{NEURIPS2021_445e24b5}.

\section{Preliminaries}\label{sec:Background}
Here, we review the setting of {\bf Bayesian learning} used by \cite{xu2020minimum} and its extension to the meta-learning setting proposed by \cite{jose2021information}. Capital letters such as $X$ represent random variables, whereas lowercase letters such as $x$ represent deterministic values.
\subsection{Bayesian Learning}\label{sec:Bayes_SBackground}
 We consider a supervised setting and denote input--output pairs by $Z=(X,Y)\in\mathcal{Z}:=\mathcal{X}\times\mathcal{Y}$. Learners can access $N$ training data, $Z^N:=(Z_1,\dots,Z_N)$ with $Z_n:=(X_n,Y_n)$, which are independent and identically distributed (i.i.d.) samples from some underlying distribution. The goal of supervised learning is to use $Z^N$ to predict the test target variable $Y$ given the test input $X$, independently drawn from the same distribution as the training data.
 For this purpose, we consider a parametric generative model. We assume that the underlying distribution belongs to a model class $\{p(z|w):w\in\mathcal{W}\}$ with model parameter $w$ in the set  $\mathcal{W}$. As pointed out by \cite{jose2021information}, \cite{hafez2021rate}, and \cite{knoblauch2019generalized}, this setting implies that the model class is well-specified. In this paper, we assume that $p(Z|W)=p(Y|X, W)p(X)$ for simplicity. This means that the input data $X$ is independent of the model parameter. 
 
 In {\bf Bayesian learning}, model parameters are treated as latent random variables following a prior distribution $p(w)$. Conditioned on $W=w$, the data are generated by $p(Z|W=w)$. Thus, the joint distribution of the training data $Z^N$, the test data $Z$, and the model parameter $W$ is given by
 \begin{align}\label{generating}
p(W,Z^N,Z)\coloneqq p(W) p(Z|W)^{N} p(Z|W).
\end{align}
Since the training data points are i.i.d. samples, we can express $p(Z|W)^N=p(Z^N|W)$. We express the Bayesian posterior as $p(W|Z^N)$ and the posterior predictive distribution as $p(Y|X,Z^N):=\mathbb{E}_{p(W|Z^N)}p(Y|X,W)$.

 Next, we introduce action $a$ and loss function $l$ to measure the performance of supervised learning. We define $\mathcal{A}$ as an action space and the loss function as $l:\mathcal{Y}\times \mathcal{A}\to\mathbb{R}$. The loss of action $a\in\mathcal{A}$ and the target variable $y$ are written as $l(y,a)$; for example, the log loss is given as $l(y,q)=-\ln q(y)$, where $q$ is the probability density of $Y$ and $\mathcal{A}$ is the set of all probability densities on $Y$. The squared loss is given as $l(y, a)=|y-a|^2$, where $\mathcal{Y}=\mathcal{A}=\mathbb{R}$. Our goal is to infer the decision rule $\psi:\mathcal{X}\times \mathcal{Z}^N\to \mathcal{A}$ that minimizes the expected loss $\mathbb{E}_{p(Y,X,Z^N)}[l(Y,\psi(X,Z^N))]$ among all decision rules. Following the previous work by \cite{xu2020minimum}, we define the infimum of the expected loss as the {\bf Bayesian risk}:
\begin{align}\label{bayesian_risk}
    R_l(Y|X,Z^N)\coloneqq \inf_{\psi:\mathcal{X}\times \mathcal{Z}^N\!\to\! \mathcal{A}}\mathbb{E}_{p(Y,X,Z^N)}[l(Y,\psi(X,Z^N))].
\end{align}
For example, when the log loss is used, $R_{\log}(Y|X,Z^N)=H[Y|X,Z^N]$, where $H[Y|X,Z^N]$ is the entropy of the posterior predictive distribution defined as
\begin{align}
H[Y|X,Z^N]=\mathbb{E}_{p(Z^N)p(X)}\mathbb{E}_{p(Y|X,Z^N)}[-\log p(Y|X,Z^N)].
\end{align}
Thus, the Bayesian risk equals the test error under a posterior predictive distribution.

Next, we define a fundamental limit of learning as $\phi:\mathcal{X}\times\mathcal{W}\to\mathcal{A}$, which takes the true parameter $W$ instead of the training dataset $Z^N$. Then, the corresponding risk is given as
\begin{align}\label{minimum}
    \scalebox{0.95}{$\displaystyle R_l(Y|X,W):=\inf_{\phi:\mathcal{X}\times\mathcal{W}\to\mathcal{A}}\mathbb{E}_{p(Y,X,W)}[l(Y,\phi(X,W))]$}.
\end{align}
We cannot improve this risk by increasing the number of training data. Thus, $R_l(Y|X,W)$ can be regarded as the {\bf aleatoric uncertainty} since it expresses the fundamental difficulty of  learning. In other words, this risk implies the inherent presence of randomness in the data-generating mechanism. When the log loss is used, $R_l(Y|X,W)$ corresponds to the conditional entropy 
\begin{align}
R_{\mathrm{log}}(Y|X,W)&=\mathbb{E}_{p(X)p(W)}H[Y|X,W]=\mathbb{E}_{p(Y,X,W)}[-\log p(Y|X,W)].    
\end{align}

Finally, we define the difference between the Bayesian risk and the fundamental limit of learning as the {\bf minimum excess risk (MER)}:
\begin{align}
    \mathrm{MER}_l(Y|X,Z^N):=R_l(Y|X,Z^N)- R_l(Y|X,W).
\end{align}
This corresponds to the {\bf epistemic uncertainty} since it is defined as the difference between the Bayesian risk and the fundamental limit of learning. Thus, MER implies the loss due to insufficient training data under the well-specified model assumption \citep{xu2020minimum,hafez2021rate}. When the log loss is used, MER  is given as
\begin{align}\label{eq_CMI}
 \mathrm{MER}_\mathrm{\log}(Y|X,Z^N)=I(W;Y|X,Z^N),
\end{align}
where $I(W;Y|X,Z^N)$ is the conditional mutual information (CMI).
Other than the log loss, if the loss function satisfies the $\sigma^2$ sub-Gaussian property conditioned on $(X,Z^N)=(x,z^N)$, \cite{xu2020minimum} showed that 
\begin{align}
\mathrm{MER}_l(Y|X,Z^N)\leq\sqrt{2\sigma^2 I(W;Y|X,Z^N)}. 
\end{align}
Thus, MER is upper-bounded by the square root of the CMI. Thus, understanding the CMI is crucial to understand $\mathrm{MER}_l$. For this reason, we focus on the log loss in this paper.

MER has some desirable properties as the epistemic uncertainty. \cite{xu2020minimum} proved that $\mathrm{MER}_l\geq 0$ and it decreases as we increase $N$. Moreover, they showed that $\mathrm{MER}_l$ can be upper-bounded by the mutual information (MI) as follows:
\begin{lemma}[\cite{xu2020minimum}]\label{lmm_existing1}
Under the joint distribution of Eq.~\eqref{generating}, we obtain
\begin{align}\label{eq_CMI_MI}
    I(W;Y|X,Z^N)\leq \frac{1}{N}I(W;Z^N).
\end{align}
\end{lemma}
In many practical settings, $I(W;Z^N)$ is upper-bounded by $\mathcal{O}(\log N)$; thus, the CMI is bounded by $\mathcal{O}(\log N/N)$. Therefore, it converges to 0 as $\mathcal{O}(\log N/N)$ for the log loss and $\mathcal{O}(\sqrt{\log N/N})$ for sub-Gaussian loss functions. It has been discussed that $I(W;Z^N)$ captures the sensitivity of the learned parameter and training dataset and is closely connected to the generalization error bound \citep{xu2017information}.

\subsection{Bayesian Meta-learning}
In traditional Bayesian inference, the prior distribution is selected on the basis of prior knowledge about the task. When we specify the appropriate prior distribution for the given task, we expect that we can reduce the number of training data we need to meet accuracy requirements. In a Bayesian meta-learning setting, the prior distribution is automatically inferred by observing related tasks. We model the statistical relationship between different tasks using a hierarchical Bayesian model with a global latent variable $U$ in the set $\mathcal{U}$. 

We observe $M$ related tasks and aim to infer a suitable prior distribution for a new unknown task. Each meta-training dataset has $N$ data points drawn i.i.d from $p(Z|W=w_m)$, where $w_m$ is the task-specific parameter.  We express the $m$-th meta-training dataset as $Z^{N,(m)}=(Z_1^{(m)},\dots,Z_N^{(m)})$. We assume that the parameter $W_m$ is drawn i.i.d from the shared prior $p(W|U)$ parametrized by the global latent variable $U$. We assume the hyperprior distribution $p(U)$ on $U$. We express the meta-training dataset as $Z^{NM}=(Z^{N,(1)},\dots,Z^{N,(M)})$. We express the model parameters of the meta-training dataset as $W^M=(W_1,\cdots,W_M)$. Finally, we have a new unknown task called the meta-test task generated using the meta-test task parameter $W$. We assume that we can use the meta-test training data $Z^N=(Z_1,\cdots, Z_N)$ and the meta-test input data $X$. 

With these settings, \cite{jose2021information} analyzed the meta-learning in Bayesian learning introduced in Sec.~\ref{sec:Bayes_SBackground}. We consider the following joint distribution:
\begin{align}\label{meta_learning}
    p(U,W^M,Z^{NM},W,Z^N,Z):=\scalebox{0.95}{$\displaystyle p(U)\underbrace{\Big(p(W|U)p(Z^N|W)\Big)^M}_{\text{meta-training}}\underbrace{p(W|U)p(Z^N|W)p(Z|W)}_{\text{meta-testing}}.$}
\end{align}
 Here, we omit the index for the meta-training dataset for simplicity, see Supplementary Material for details. Under this setting, we consider the decision rules and excess risk in the same way as in  Sec.~\ref{sec:Bayes_SBackground}. We define the Bayesian meta-risk as
\begin{align}
&R_l(Y|X,Z^N,Z^{NM})\coloneqq \inf_{\psi_{\mathrm{meta}}:\mathcal{X}\times \mathcal{Z}^{NM}\times\mathcal{Z}^N\!\to\! \mathcal{A}}\mathbb{E}_{p(Z^{NM},Z^N,Z)}[l(Y,\psi_{\mathrm{meta}}(X,Z^{NM},Z^N))].
\end{align}
We also define the fundamental limit of learning in meta-learning as
\begin{align}
&R_l(Y|X,W,U)\displaystyle:=\inf_{\phi_{\mathrm{meta}}:\mathcal{X}\times\mathcal{W}\times\mathcal{U}\to\mathcal{A}}\mathbb{E}_{p(U,W,Z)}[l(Y,\phi_{\mathrm{meta}}(X,W,U))].
\end{align}
We then define the minimum excess meta-risk (MEMR) as
\begin{align}
        \mathrm{MEMR}_l(Y|X,Z^N,Z^{NM}):=R_l(Y|X,Z^N,Z^{NM})- R_l(Y|X,W,U).
\end{align}
\cite{jose2021information} showed that the MEMR of the log loss equals to the CMI:
\begin{align}
    \mathrm{MEMR}_{\log}(Y|X,Z^N,Z^{NM})=I(Y;W|X,Z^N,Z^{NM}),
\end{align}
and derived the upper-bound of $\mathrm{MEMR}_{\log}$, which is similar to Eq.~\eqref{eq_CMI_MI}, as
\begin{align}\label{eq_memr}
    \scalebox{0.92}{$\displaystyle I(Y;W|X\!,Z^N\!,Z^{NM})\leq  \frac{I(U;Z^{NM})}{NM}\!+\frac{I(W;Z^N|U)}{N}.$}
\end{align}
We can see that the CMI is also upper-bounded by the MI that captures the sensitivities of the learned meta-test task parameter, hyperparameter, and meta-training dataset.

\section{Exact Characterization of CMI}\label{sec:exact}
Here, we show our novel CMI decomposition and present the information-theoretic quantity of the sensitivity. First, we consider the Bayesian learning setting introduced in Sec.~\ref{sec:Bayes_SBackground}. All the proofs are shown in Supplementary Material.

\subsection{Information-theoretic Decomposition of CMI}
As we pointed out in Sec.~\ref{sec:intro}, the analysis of $\mathrm{MER}_l$ introduced in Sec.~\ref{sec:Background} cannot explain the sensitivity in uncertainty between the test and training data. The limitation of the information-theoretic analysis of Lemma~\ref{lmm_existing1} is that the decomposition of the CMI focuses only on the training dataset and not on the test data point.

The following theorem is our first main result, which decomposes the CMI as the sum of the MI and the sensitivities of the test data and each training data point.
\begin{theorem}\label{thm_main_cmd}
Under the joint distribution of Eq.~\eqref{generating}, we have
\begin{align}\label{thm_main_cmd_eq}
    \scalebox{0.97}{$\displaystyle I(W;Y|X,Z^N)\!=\!\frac{1}{N}I(W;Z^N)\!-\!\frac{1}{N}\sum_{n=1}^N I(Z,Z_n|Z^{N\backslash n})\!-\!\frac{1}{N}\sum_{n'=1}^{N-1}\sum_{n=n'}^{N-1}I(Z_{n+1},Z_{n}|Z^{n-1})$},
\end{align}
where $Z^{N \backslash n}:=(Z_1,\dots,Z_{n-1},Z_{n+1},\dots,Z_N)$ and $Z^{n-1}:=(Z_1,\dots,Z_{n-1})$.
\end{theorem}
Different from the bound in Lemma~\ref{lmm_existing1}, the CMI is decomposed into three terms connected with equality, not inequality.
The first term on the right-hand side of Eq.~\eqref{thm_main_cmd_eq} is the MI between the learned parameter and the training dataset.  The second and third terms correspond to the binary relation about how much information each data point has to predict other data points.  The second term $I(Z,Z_n|Z^{N\backslash n})$ represents the information-theoretic quantity of the sensitivity of the test and training data points. This term indicates how useful the training data point $Z_n$ is to predict the test data point $Z$. If the training data point $Z_n$ has more information about the test data, then the uncertainty at $Z$ decreases. If $Z_n$ is almost independent of $Z$, then the mutual information becomes $0$, which means that the uncertainty increases. 

From this observation, we introduce the definition of test data sensitivity as follows.
\begin{definition}The sensitivity of the test data and $n$-th training data point is defined as
\begin{align}
I_{n}\coloneqq I(Z,Z_n|Z^{N\backslash n}).
\end{align}
\end{definition}
For simplicity, we also express $I_{n+1,n}:=I(Z_{n+1},Z_{n}|Z^{n-1})$, which appears in Eq.~\eqref{thm_main_cmd_eq}. 

We note that since $X$ and $W$ are independent of each other, $I(Z,Z_n|Z^{N\backslash n})=I(Y,Y_n|X,X_n,Z^{N\backslash n})$ holds. We can transform $I_n$ into a more intuitive expression as
\begin{align}
    I_n&=H(Z|Z^{N\backslash n})-H(Z|Z^N) \label{entropy_diff}\\
    &\scalebox{0.95}{$\displaystyle=\mathbb{E}_{p(W,Z^N,Z)}\ln\frac{\mathbb{E}_{p(W|Z^{N \backslash n})}p(Z,Z_{n}|W)}{p(Z|Z^{N \backslash n})p(Z_n|Z^{N \backslash n})}\label{joint_pred}$}.
\end{align}
Eq.~\eqref{entropy_diff} is useful for explicitly calculating the sensitivity for some models shown in Sec.~\ref{sec_LR}. Eq.~\eqref{joint_pred} states that the joint posterior predictive distribution $\mathbb{E}_{p(W|Z^{N \backslash n})}p(Z,Z_{n}|W)$ differs from the single-point posterior predictive distribution. The joint predictive distribution has recently attracted attention in decision problems \citep{rosenfeld2020predictions,osband2021evaluating,https://doi.org/10.48550/arxiv.2107.09224}. Thus, our theoretical results suggest new insights into the connection between decision problems, joint predictive distribution, and uncertainty. However, this is outside the scope of this study, and we leave it to future work to explore this connection.

Finally, from Theorem~\ref{thm_main_cmd}, we obtain the new information-theoretic bound for the CMI as follows.
\begin{corollary}\label{col_improved_bound}
Under the joint distribution of Eq.~\eqref{generating}, we obtain
\begin{align}
    I(W;Y|X,Z^N)\displaystyle\leq \frac{1}{N}I(W;Z^N)-\frac{1}{N}\sum_{n'=1}^{N-1}\sum_{n=n'}^{N-1}I(Z_{n+1},Z_{n}|Z^{n-1}).
\end{align}
\end{corollary}
 This bound is tighter than that of Lemma~\ref{lmm_existing1} owing to the second term on the right-hand side. In Sec.~\ref{sec_exp}, we numerically compare this bound with that of Lemma~\ref{lmm_existing1}.

\subsection{Linear Regression Model}\label{sec_LR}
In this section, we use a linear regression model to explore the sensitivity between the test and training data. The likelihood of the model is given as the Gaussian distribution with the mean $w^\top \phi(x)$ and the variance $\beta^{-1}\in\mathbb{R}^+$. We express it as $p(y|x,w)=\mathcal{N}(w^\top \phi(x),\beta^{-1})$, where $\mathcal{Y}=\mathbb{R}$ and $\phi(x):=(\phi_1(x),\dots,\phi_d(x))^\top\in\mathbb{R}^d$ is a $d$-dimensional feature vector of the input $x$ and each $\phi_i:\mathcal{X}\to\mathbb{R}$. We assume a prior distribution $p(w)=\mathcal{N}(0,\alpha^{-1} I_d)$ with some positive constant $\alpha>0$. We define a design matrix as $\Phi=(\phi(x_1),\dots,\phi(x_N))^\top\in\mathbb{R}^{N\times d}$. We also define a target vector as $\mathbf{y}=(y_1,\dots,y_N)^\top$. Then, a posterior distribution is given by $p(w|z^N)=\mathcal{N}(m_N,S_N)$, where $m_N=\beta S_N\Phi^\top \mathbf{y}$ and $S_N^{-1}:=\alpha I_d+\beta \Phi^\top\Phi$. We also have a posterior predictive distribution as $p(y|x,z^N):=\mathcal{N}(m_N^\top\phi(x),\sigma^2_N(x))$, where $\sigma^2_N(x):=\beta^{-1}+\phi(x)^\top S_N\phi(x)$. 

Since the posterior predictive distribution is given as the Gaussian distribution, its entropy is calculated on the basis of its variance. Thus, $I(W;Y|X,Z^N)=\mathbb{E}_{p(X)}\log \sigma_N^2(X)/2+\mathrm{Const}$, and the interplay between $\phi(x)$ and $S_N$ characterizes the sensitivity of the test and training data points. Similar arguments still hold for Gaussian process models, where the inner products of the feature maps are replaced with kernel functions.

We can explicitly calculate the sensitivity $I_n$  using Eq.~\eqref{entropy_diff}. Then, we obtain
\begin{align}
&I_n=\mathbb{E}_{p(X^{N+1})}\frac{1}{2}(\ln \sigma^2_{N\backslash n}(X)-\ln \sigma^2_N(X) ).
\end{align}
We can simplify this as follows.
\begin{theorem}\label{LM_bounds}For linear models, the sensitivity $I_n$ satisfies the following relation:
\begin{align}
    \mathbb{E}_{p(X^{N+1})}\frac{(\phi(X)^\top S_N\phi(X_n))^2}{2\omega(X_n)(\alpha^{-1}\phi(X)^\top\phi(X)+\beta^{-1})}\leq I_n\leq \mathbb{E}_{p(X^{N+1})}\frac{(\phi(X)^\top S_N\phi(X_n))^2}{2\omega(X_n)(\beta^{-1}+\phi(X)^\top S_N\phi(X)},\nonumber
\end{align}
where $\omega(x):=\beta^{-1}-\phi(x)^\top S_N\phi(x)$.
\end{theorem}
This bound implies that the posterior covariance matrix $S_N:=(\alpha I_d+\beta \Phi^\top\Phi)^{-1}$ can be seen as a metric for measuring the similarity between the training data $x_n$ and the test data $x$. In Supplementary Material, we numerically evaluated this bound. 

Combined with Theorem~\ref{thm_main_cmd}, we obtain
\begin{align}
    &\mathrm{MER}_{\log}(Y|X,Z^N)\leq \frac{1}{N}I(W;Z^N)-\mathbb{E}_{p(X^{N+1})}\frac{1}{N}\sum_{n=1}^N\frac{(\phi(X)^\top S_N\phi(X_n))^2}{2\omega(X_n)(\alpha^{-1}\phi(X)^\top\phi(X)+\beta^{-1})}\nonumber.
\end{align}
This suggests that the test error becomes small if the given test and training data points are similar under the feature map with the metric $S_N$.

\subsection{Asymptotic Behavior}
Here, we discuss the asymptotic behavior of sensitivity. Using the asymptotic expansion of Bayesian inference introduced in \cite{watanabe2018mathematical}, we obtain the following relation:
\begin{theorem}\label{thm_asymp2}
Assume that $p(z|w)$ has a relatively finite variance, that is, for any pair of $w_0, w\in\mathcal{W}$, there exists a positive constant $c_0$ such that
\begin{align}\label{ass_rela}
    c_0\mathbb{E}_{p(Z|w_0)}(\ln p(Z|w_0)-\ln p(Z|w))^2\leq \mathbb{E}_{p(Z|w_0)}[\ln p(Z|w_0)-\ln p(Z|w)].
\end{align}
Then, we obtain
    $\displaystyle I_n=I(Z,Z_n|Z^{N\backslash n})=o\left(\frac{1}{N}\right)$,
where $o(\frac{1}{N})$ is little $o$.
\end{theorem}
The relatively finite variance assumption in Eq.~\eqref{ass_rela} is satisfied in many widely used models. For example, generalized linear models, including the linear and logistic regression models, satisfy this condition. See Supplementary Material and \cite{watanabe2018mathematical} for other examples.

Combined with Theorem~\ref{thm_main_cmd}, since $\frac{1}{N}I(W;Z^N)=O\left(\frac{1}{N}\right)$, we obtain
\begin{align}\label{eq_order_info}
    I(W;Y|X,Z^N)&\leq \underset{=O\left(\frac{1}{N}\right)}{\underline{\frac{1}{N}I(W;Z^N)}}-\underset{=o\left(\frac{1}{N}\right)}{\underline{\frac{1}{N}\sum_{n=1}^N I(Z,Z_n|Z^{N\backslash n})}}.
\end{align}
Thus, since the order of the sensitivity term is $o(1/N)$, it is much smaller than the MI, which is $O(1/N)$. Finally, using Theorem~\ref{thm_asymp2}, we obtain the following relation:
\begin{align}\label{eq_val}
    &\frac{1}{N}\sum_{n'=1}^{N-1}\sum_{n=n'}^{N-1}I(Z_{n'+1},Z_{n'}|Z^{n'-1})=O\left(\frac{1}{N}\right).
\end{align}

\section{Exact Characterization of Generalization Error}
Here, we present the application of Theorem~\ref{thm_main_cmd} to the generalization error analysis. 

\subsection{Relation to Generalization Error}\label{bayes}
First, we show that Lemma~\ref{lmm_existing1} is closely related to the generalization error.
\begin{lemma}\label{lem_cd_MI}
When a log loss is used, $I(W;Y|X,Z^N)\leq \frac{1}{N}I(W;Z^N)$ is equivalent to the following inequality;
\begin{align}\label{eq_CMI_MI_test}
R_{\log}(Y|X,Z^N) &\leq -\mathbb{E}_{p(Z^N)}\ln p(Z^N)\notag\\
&= -\mathbb{E}_{p(Z^N)}\mathbb{E}_{p(W|Z^N)}\frac{1}{N}\sum_{n=1}^N\ln p(Z_n|W)+\frac{1}{N}\mathrm{KL}(p(W|Z^N)|p(W)).
\end{align}
\end{lemma}
The left-hand side of Eq.~\eqref{eq_CMI_MI_test} is the test error, and the right-hand side is the training error plus the regularization term. Thus, Lemma~\ref{lmm_existing1} is closely related to the generalization error.  With this observation, using Theorem~\ref{thm_main_cmd}, we can incorporate the sensitivity of the test and training data to the generalization error as follows.
\begin{theorem}\label{eq_bayesian_generalization}
Under the joint distribution of Eq.~\eqref{generating} with a log loss, we obtain
\begin{align}
R_{\log}(Y|X,Z^N)
    &=-\mathbb{E}_{p(Z^N)}\mathbb{E}_{p(W|Z^N)}\frac{1}{N}\sum_{n=1}^N\ln p(Y_n|X_n,W)+\frac{1}{N}\mathrm{KL}(p(W|Z^N)|p(W))\notag \\
    &\quad-\frac{1}{N}\sum_{n=1}^N I_n-\frac{1}{N}\sum_{n'=1}^{N-1}\sum_{n=n'}^{N-1}I_{n+1,n}\label{colleration}.
\end{align}
\end{theorem}
From Eq.~\eqref{colleration}, if the training data $x_n$ has sufficient information to predict $x$, $I_n$ becomes large, leading to a smaller test error. Thus, this relation formalizes our intuition that we can predict a test data point, which is similar to the training data in some sense, better than the test data, which are completely different from the training data.

Another interesting point is that, unlike Lemma~\ref{lem_cd_MI}, Eq.~\eqref{colleration} is the identity, not the inequality. Thus, we can precisely characterize the relationship between the test and training errors. We will discuss the relation between our result and the recently proposed exact characterization of the generalization error \citep{NEURIPS2021_445e24b5} in Sec.~\ref{exact_sim}.

\subsection{Relationship between the Sensitivity and the Gibbs Test Error}\label{exact_sim}
In many generalization error analyses, we often use the Gibbs test error defined as,
\begin{align}
    R_{\log}^{\mathrm{Gibbs}}(Y|X,Z^N):=\scalebox{0.925}{$\displaystyle\mathbb{E}_{p(W)p(Z^N|W)p(\tilde{W}|Z^N)}[-\mathbb{E}_{p(Z|W)}\log p(Y|X,\tilde{W})]$}.
\end{align}
 Here, we express the learned parameter as $\tilde{W}$, which follows the Bayesian posterior distribution $p(\tilde{W}|Z^N)$. 
 By comparing with $R_{\log}(Y|X,Z^N)$, which uses the posterior predictive distribution, we obtain
\begin{align}\label{eq_jensen}
    &R_{\log}(Y|X,Z^N)\leq R_{\log}^{\mathrm{Gibbs}}(Y|X,Z^N),
\end{align}
where we used the Jensen inequality. This relation is general since we only use the convexity of the log loss. 

We further explore the relationship between the Gibbs test error $R_{\log}^{\mathrm{Gibbs}}(Y|X,Z^N)$ and the Bayesian risk $R_{\log}(Y|X,Z^N)$ using the Lautum information (LI), which was used by \cite{NEURIPS2021_445e24b5}.
First, we present the exact characterization of the generalization error of the Gibbs test error.
\begin{theorem}\label{inside_term}
Under the joint distribution of Eq.~\eqref{generating} with a log loss, we obtain
\begin{align}\label{eq_gene_LI}
    \scalebox{0.94}{$\displaystyle R_{\log}^{\mathrm{Gibbs}}$} (Y| X,Z^N)&\scalebox{0.94}{$\displaystyle=-\mathbb{E}_{p(W)p(Z^N|W)}\frac{1}{N}\sum_{n=1}^N\ln p(Y_n|X_n,W)$}+\frac{1}{N}LI(\tilde{W};Z^N|W),
\end{align}
where LI is the conditional Lautum information defined as
\begin{align}
LI(\tilde{W};Z^N|W)&=\mathbb{E}_{p(W)p(\tilde{Z}^N|W)p(\tilde{W}|\tilde{Z}^N)p(Z^N|W)}\log \frac{p(Z^N|W)p(\tilde{W}|W)}{p(Z^N,\tilde{W}|W)}\nonumber \\
&=\mathrm{KL}(p(\tilde{W}|W)p(Z^N|W)|p(\tilde{W},Z^N|W)).
\end{align}
\end{theorem}
Note that the LI is closely related to the reverse KL divergence \citep{NEURIPS2021_445e24b5}.
In conclusion, the generalization error of the Gibbs test error is characterized by the LI. This result is similar to the exact characterization of generalization error under the {\bf frequentist setting} used by \cite{NEURIPS2021_445e24b5}. They assumed that the data is drawn i.i.d from an unknown distribution and the model's parameters are not treated as latent variables. Then, they proved that the generalization error between the Gibbs test and training errors is equivalent to the sum of the LI and the MI between learned parameters and training datasets.

Combining Theorems~\ref{eq_bayesian_generalization} and \ref{inside_term}, we obtain the following exact characterization of the Jensen gap:
\begin{corollary}
Under the joint distribution of Eq.~\eqref{generating}, we obtain
\begin{align}\label{jensen_gap}
    \scalebox{0.943}{$\displaystyle R_{\log}^{\mathrm{Gibbs}}(Y|X,Z^N)\!-\!R_{\log}(Y|X,Z^N)\!=\!\frac{LI(\tilde{W};Z^N|W)}{N}\!+\!\sum_{n=1}^N \frac{I_n}{N}\!+\!\sum_{n'=1}^{N-1}\!\sum_{n=n'}^{N-1}\frac{I_{n+1,n}}{N}\!-\!\frac{I(W;\!Z^N)}{N}.$}
\end{align}
\end{corollary}
From the Jensen inequality of Eq.~\eqref{eq_jensen}, if the posterior $p(\tilde{W}|Z^N)$ is a point mass, the Jensen gap vanishes. From this relation, as the sensitivity term $I_n$ increases, the Jensen gap becomes large. We point out that the Jensen gap has been studied in relation to the model misspecification under the frequentist setting \citep{grunwald2012safe,grunwald2017inconsistency}. Since our setting is Bayesian learning, it is difficult to directly compare our Eq.~\eqref{jensen_gap} with previously reported results of the frequentist setting. We leave it to future work to clarify how the existing analysis of the Jensen gap under model misspecification is translated into our setting.

\section{Exact Characterization of CMI in Meta-learning}\label{sec_meta}
In this section, we extend our information-theoretic analysis of the sensitivity to a Bayesian meta-learning setting. The following is our main result:
\begin{theorem}\label{thm_main_cmd_meta}
Under the joint distribution of Eq.~\eqref{meta_learning}, we obtain
\begin{align}
    I(Y;W|X,Z^N,Z^{NM})
    &=\frac{1}{N}I(W;Z^N|U)+\frac{1}{NM}I(U;Z^{NM})\label{eq_task_memer} \\
    &-\frac{1}{NM}\sum_{m=1}^MI(Z^N,Z^{N,(m)}|Z^{N(M\backslash m)})\label{eq_task}\\
    &-\frac{1}{NM}\sum_{m'=1}^{M-1}\sum_{m=m'}^{M-1}I(Z^{N,(m+1)},Z^{N,(m)}|Z^{N(m-1)})\label{eq_task2}\\
    &-\frac{1}{N}\sum_{n=1}^N I(Z,Z_n|Z^{N\backslash n},Z^{NM})\label{eq_data}\\
    &-\frac{1}{N}\sum_{n'=1}^{N-1}\sum_{n=n'}^{N-1}I(Z_{n+1},Z_{n}|Z^{n-1},Z^{NM})\label{eq_data2},
\end{align}
where  $Z^{Nm}:=(Z^{N,(1)},\cdots,Z^{N,(m)})$, $Z^{N(m-1)}:=(Z^{N,(1)},\cdots,Z^{N,(m-1)})$, and \\$Z^{N(M\backslash m)}:=(Z^{N,(1)},\dots,Z^{N,(m-1)},Z^{N,(m+1)},\dots,Z^{N,(M)})$.
\end{theorem}
Compared with the existing bound in Eq.~\eqref{eq_memr}, the terms of Eqs.~\eqref{eq_task} to \eqref{eq_data2} newly appeared. Eq.~\eqref{eq_task} represents the sensitivity between the test and training tasks since it quantifies how useful the $m$-th training task is to predict the meta-test task. To the best of our knowledge, our study is the first to theoretically quantify task sensitivity. Eq.~\eqref{eq_data} quantifies the sensitivities of the meta-test training data and meta-test test data points similarly to Theorem~\ref{thm_main_cmd}. 

We can evaluate information-theoretic quantities by considering the following posterior and predictive distributions:
\begin{align}
p(Y|X,Z^N,Z^{NM})=\mathbb{E}_{p(W|Z^N,Z^{NM})}p(Y|X,W)=\mathbb{E}_{p(U|Z^{NM})p(W|Z^N,U)} p(Y|X,W)\nonumber 
\end{align}
The information from the relevant tasks is captured by the hyper-posterior distribution $p(U|Z^{NM})$, and the information from meta-test training data is incorporated into the posterior distribution $p(W|Z^N,U)$. These correspond to Eq.~\eqref{eq_task_memer}, which also appears in the existing MEMR bound. The sensitivity between the meta-test and meta-training tasks of Eq.~\eqref{eq_task} is given as
\begin{align}
    \scalebox{0.95}{$\displaystyle I(Z^N,Z^{N,m}|Z^{N(M\backslash m)})\!=\!H(Z^N|Z^{N(M\backslash m)})\!-\!H(Z^N\!|Z^{NM})$}.
\end{align}
We can evaluate the left-hand side by evaluating the hyper-posterior distributions $p(U|Z^{N(M\backslash m)})$ and $p(U|Z^{NM})$. Note that we can obtain the improved information-theoretic upper bound about MEMR, which improves Eq.~\eqref{eq_memr}, in the same way as Corollary~\ref{col_improved_bound}.

\section{Related Work}
The sensitivity between test and training data points has been an important property theoretically and practically \citep{PRML,murphy2012machine}. Linear models and Gaussian processes have extensively been studied to analyze the sensitivity since their posterior predictive distribution is expressed analytically \citep{fiedler2021practical,lederer2019posterior}. Our result extends this relationship to general probabilistic models using the information-theoretic quantity for the first time. Such a relationship provides an important contribution in practice since some recent studies, such as \cite{angelopoulosuncertainty,liu2020simple,tian2021geometric}, and \cite{he2020bayesian},  explicitly introduced the sensitivity property into deep neural networks to enhance the uncertainty quantification performance. Similarly to sensitivity, the CMI is widely used as the objective function in Bayesian experimental designs \citep{foster2019variational}. Thus, understanding the sensitivity of the CMI will lead to the analysis of such applications. In meta-learning tasks, information-theoretic quantities are widely used \citep{pmlr-v161-titsias21a, pmlr-v151-chen22h} to quantify the similarity of tasks. In these applications, evaluating exact information-theoretic quantities is difficult for many practical models. Thus, various approximation methods have been proposed, including variational inference \citep{PRML}. We leave it to future work to explore how the approximation quality affects the sensitivity in uncertainty.

The information-theoretic analysis has recently received attention in the generalization error analysis \citep{xu2017information, pensia2018generalization}.
In such generalization error analysis, including the PAC-Bayesian theory \citep{NIPS2017_7edccc66,alquier2021user}, the data generating distribution may not be well-specified, and model parameters are not treated as latent variables. Compared with previous studies, we can specify the correct model families in the Bayesian learning settings. Finally, we note that the joint model in Eq.~\eqref{generating} often appears in the Bayesian decision theory \citep{robert2007bayesian} in statistics. This model evaluates the {\it average} performance of the risk function over a prior distribution and leads to the minimax rate analysis of the parameter estimation. The lower bound for the decision rule in the minimum excess risk has recently been reported \citep{hafez2021rate} using the rate-distortion theory \citep{10.5555/1146355}. Moreover, the joint model in Eq.~\eqref{generating} is used in Bayesian experimental design, in which the stochastic dependencies of data and parameters are introduced to incorporate uncertainty. We also remark that this model is closely related to Bayesian online learning, and we show the regret analysis in Supplementary Material.

\section{Numerical Experiments}\label{sec_exp}
Here, we show the numerical evaluation of the sensitivities in Theorems~\ref{thm_main_cmd} and \ref{thm_main_cmd_meta}. Detailed experimental settings and additional results are shown in Supplementary Material.

\subsection{Experiments in Bayesian Learning Setting}\label{sec_not_meta_exp}
First, using linear regression models introduced in Sec.~\ref{sec_LR}, we numerically evaluated information-theoretic quantities appearing in Theorem~\ref{thm_main_cmd}, changing the training data size $N$. Note that we can calculate all the information-theoretic quantities analytically. In the main paper, we only show the results of Gaussian basis functions as feature map $\phi$, whose dimension is set to $10$. 

\begin{figure}[tb!]
 \centering
 \includegraphics[width=0.9\linewidth]{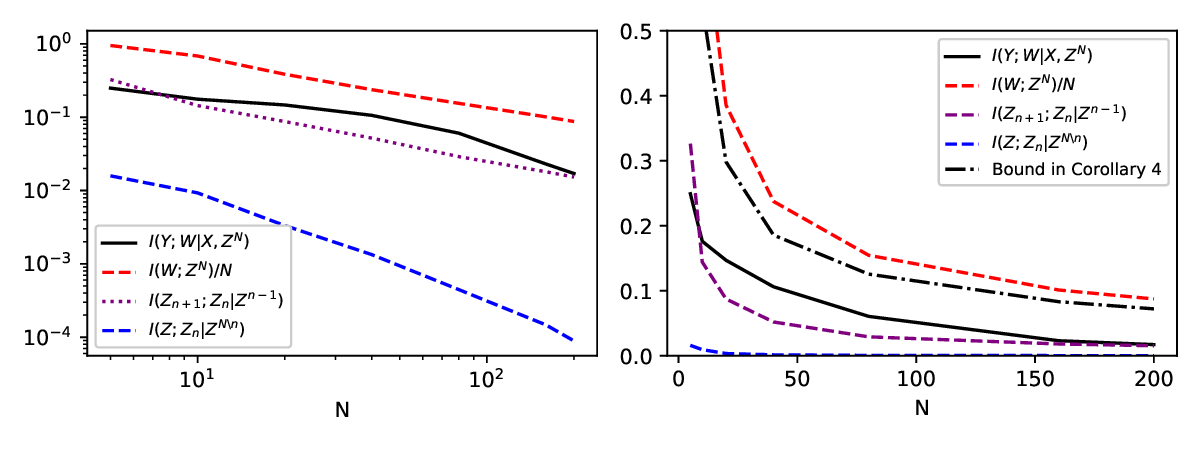}
 \vspace{-0.9truemm}
 \caption{Information-theoretic quantities appearing in Theorem~\ref{thm_main_cmd} and Corollary~\ref{col_improved_bound}.}
\label{fig_ampm}
\end{figure}

 The results are shown in Fig.~\ref{fig_ampm}, where we plot the CMI ($I(Y;W|X,Z^N)$), MI ($I(W;Z^N)/N$), and the sum of the test data sensitivity ($\sum_n I(Z;Z_n|Z^{(N\backslash n)})/N$) and the sum of the training data sensitivity ($\frac{1}{N}\sum_{n'=1}^{N-1}\sum_{n=n'}^{N-1}I(Z_{n+1};Z_n|Z^{(n-1)})$). In the figure legend, we omit the summation with respect to $n$ and $n'$ for clarity. In the left panel of this figure, we plot them in the log scale, and we can see that all the terms converge linearly in the plot. This is consistent with Eq.~\eqref{eq_order_info}, which describes the asymptotic order of each quantity. Note that the sensitivity term $I_n$ converges faster than the other terms, as indicated by the asymptotic analysis in Theorem~\ref{thm_asymp2}. In the right panel of this figure, in addition to the CMI, MI, and sensitivities, we plot our proposed bound in Corollary~\ref{col_improved_bound}. Our bound is tighter than the existing bound, which corresponds to $I(W;Z^N)/N$ owing to the sensitivity given as $I(Z_{n+1};Z_n|Z^{(n-1)})$. In Supplementary Material, we numerically evaluated the upper and lower bounds of $I_n$ in Theorem~\ref{thm_main_cmd}.

\subsection{Experiments in Bayesian Meta-learning Setting}
Next, we numerically evaluated the theoretical findings of meta-learning settings in Theorem~\ref{thm_main_cmd_meta}. For this purpose, we put a hyperprior on the parameters of the linear regression model. We consider that $p(W|U)=\mathcal{N}(U,\alpha^{-1}I_d)$ and $p(U)=\mathcal{N}(0,\gamma^{-1}I_d)$. Under these settings, we can analytically calculate the posterior distributions $p(U|Z^{NM}), p(W|Z^N,U)$, and $p(Y;W|X,Z^N,Z^{NM})$. Thus, we can analytically evaluate the information-theoretic quantities in Theorem~\ref{thm_main_cmd_meta}, see Supplementary Material for details. 

\begin{figure}[tb!]
 \centering
 \includegraphics[width=0.9\linewidth]{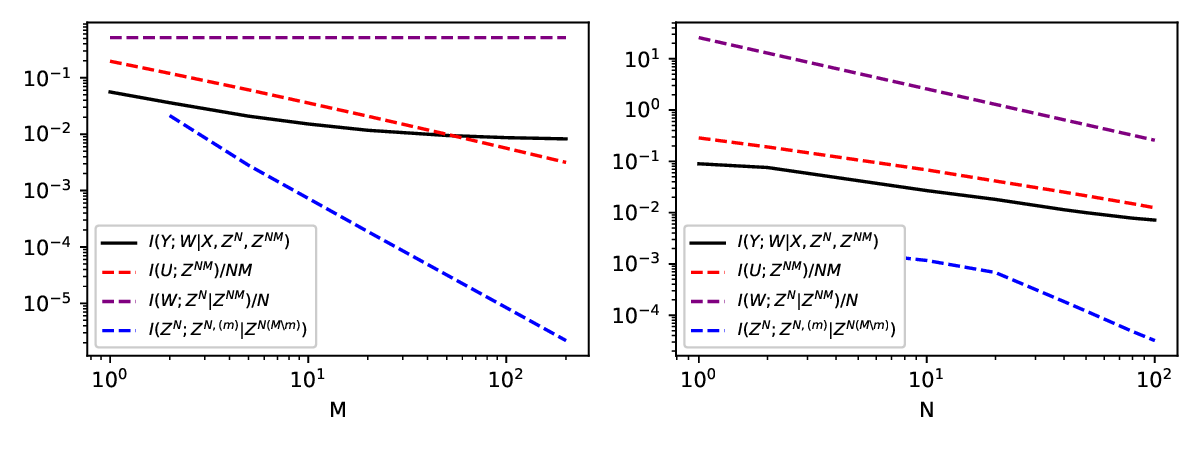}
 \vspace{-0.9truemm}
 \caption{Information-theoretic quantities appearing in Theorem~\ref{thm_main_cmd_meta}. The left panel shows the results under different $M$ (the number of tasks) at fixed $N$ (the number of training datasets), and the right panel shows the results under different $N$ at fixed $M$. In the legends, we omit the summation with respect to $n, n', m$, and $m'$ for clarity.}
\label{fig_meta}
\end{figure}

The result is shown in Fig.~\ref{fig_meta}.
In the left panel of this figure, fixing $N=50$, we plot the information-theoretic quantities with increasing $M$. We can see that MEMR (black line) decreases as we increase the number of meta-training datasets. We can also see that the MI between the hyperparameter and meta-training datasets (red dot line) also decreases. Finally, we can see that the sensitivity between the meta-test and meta-training tasks decreases faster than other information-theoretic quantities. In the right panel of this figure, fixing $M=20$, we plot the information-theoretic quantities with increasing $N$. By increasing $N$, we find that all the quantities decrease as we expected.

\section{Conclusion}
In this work, we showed the novel decomposition of the CMI and then provided the information-theoretic quantity of the sensitivity between the test and training data points. Our analysis rigorously characterizes the uncertainty's widely believed sensitivity property for the first time. Our analysis is also extended to the meta-learning setting and showed the sensitivity between tasks for the first time. It will be interesting to analyze the sensitivity under model misspecification and approximation in future work.

\bibliography{sensitivity}

\appendix

\section{Proof of Theorem~\ref{thm_main_cmd}}
\proof
First, we show the following lemma.
\begin{lemma}\label{lmm_entropuy}
Under the same setting as Theorem~\ref{thm_main_cmd}, for any $k\in (1,N]$, we have
\begin{align}
    H[Z_{k+1}|Z^{k}]=H[Z_{k+1}|Z^{k-1}]-I(Z_{k+1};Z_{k}|Z^{k-1}).
\end{align}
\end{lemma}
We can derive this by directly calculating the definition of entropy and conditional mutual information. This lemma states how much uncertainty at $Z_{k+1}$ reduces by adding a data point $Z_k$. From this relation, we get the relation about the conditional mutual information as follows,
\begin{lemma}\label{MI_basic}Under the same setting as Theorem~\ref{thm_main_cmd}, for any $k\in (1,N]$, we have
\begin{align}
    I(W;Z_{k+1}|Z^{k})=I(W;Z_k|Z^{k-1})-I(Z_{k+1};Z_{k}|Z^{k-1}).
\end{align}
\end{lemma}
\proof
From the definition of conditional mutual information, we have
\begin{align*}
I(W;Z_{k}|Z^{k-1})&=H[Z_k|Z^{k-1}]-H[Z_k|W,Z^{k-1}]\\
&=H[Z_{k+1}|Z^{k-1}]-H[Z_k|W]\\
&=H[Z_{k+1}|Z^{k}]+I(Z_{k+1},Z_k|Z^{k-1})-H[Z_{k+1}|W]\\
&=H[Z_{k+1}|Z^{k}]+I(Z_{k+1},Z_k|Z^{k-1})-H[Z_{k+1}|W,Z^K]\\
&=I[W;Z_{k+1}|Z^{k}]+I(Z_{k+1},Z_k|Z^{k-1}),
\end{align*}
where the second equality is because the joint distribution of $(Z^{k-1},Z_k)$ is equivalent to the joint distribution of $(Z^{k-1},Z_{k+1})$, and $Z^{k-1}$ is independent of $Z_k$ given $W$. The third inequality is because of Lemma~\ref{lmm_entropuy}. 
\BlackBox\\
This relation explains how much information about $W$ we obtain by adding a data point $Z_k$ into the training dataset.

Finally, by using the chain rule of the mutual information and applying Lemma~\ref{MI_basic} recursively, we obtain
\begin{align}\label{eq_derivation}
    I(W;Z^N)&=\sum_{n=1}^NI(W;Z_n|Z^{n-1})\nonumber \\
    &=NI(W;Z_N|Z^{N-1})+\sum_{n'=1}^{N-1}\sum_{n=n'}^{N-1}I(Z_{n+1},Z_{n}|Z^{N-1})
\end{align}
We transform the first term. By definition, we have
\begin{align*}
    I(W;Z_N|Z^{N-1})&=H[Z_{N}|Z^{N-1}]-H[Z_N|W]\\
    &=\mathbb{E}_{p(W)p(Z^N|W)} [-\ln p(Z^N)+\ln p(Z^{N-1})] -H[Z_n|W].
\end{align*}
Here $Z^{N-1}=(Z_1,\cdots,Z_{N-1})$. Let us consider the distribution of $Z^{N\backslash n}=(Z_1,\dots,Z_{n-1},Z_{n+1},\dots,Z_N)$. We express the marginal distribution of this as $p(Z^{N-1})=\mathbb{E}_{p(W)}p(Z^{N-1}|W)$ and $p(Z^{N\backslash n})=\mathbb{E}_{p(W)}p(Z^{N\backslash n}|W)$. Note that $p(Z^N|W)$ and $p(Z^{N\backslash n}|W)$ are equivalent since all the data is drawn i.i.d. Thus, from the chain rule of the KL divergence $\mathbb{E}_{p(W)p(Z^N|W)} [\ln p(Z^{N-1})-\ln p(Z^{N\backslash n}|W)]=0 $. Thus we have
\begin{align*}
    I(W;Z_N|Z^{N-1})&=H[Z_{N}|Z^{N-1}]-H[Z_N|W]\\
    &=\mathbb{E}_{p(W)p(Z^N|W)} [-\ln p(Z^N)+\ln p(Z^{N-1})] -H[Z_n|W]\\
    &=\mathbb{E}_{p(W)p(Z^N|W)} [-\ln p(Z^N)+\ln p(Z^{N\backslash n})] -H[Z_n|W]=I(W;Z_n|Z^{N\backslash n}).
\end{align*}
Then we substitute this to Eq.~\eqref{eq_derivation}
\begin{align}
    I(W;Z^N)&=NI(W;Z_N|Z^{N-1})+\sum_{n'=1}^{N-1}\sum_{n=n'}^{N-1}I(Z_{n+1},Z_{n}|Z^{N-1})\nonumber \\
    &=\sum_{n=1}^NI(W;Z_n|Z^{N\backslash n})+\sum_{n'=1}^{N-1}\sum_{n=n'}^{N-1}I(Z_{n+1},Z_{n}|Z^{N-1}).
\end{align}
Finally, we apply Lemma~\ref{MI_basic} to each $I(W;Z_n|Z^{N\backslash n})$. Then, for each $n\in (1, N]$ we have
\begin{align}
    I(W;Z_{N+1}|Z^{N})=I(W;Z_n|Z^{N-1})-I(Z_{N+1};Z_{n}|Z^{N-1}).
\end{align}
This is because the training dataset $\{Z_n\}_{n=1}^N$ are i.i.d., and thus, we can permute the index of the training data points in  Lemma~\ref{MI_basic}. By setting $Z_{N+1}:=Z$, which is the test data point, we obtain
\begin{align}
    I(W;Z^N)&=\sum_{n=1}^NI(W;Z_n|Z^{N\backslash n})+\sum_{n'=1}^{N-1}\sum_{n=n'}^{N-1}I(Z_{n+1},Z_{n}|Z^{N-1})\nonumber \\
    &=NI(W;Z|Z^N)+\sum_{n=1}^NI(Z,Z_n|Z^{N\backslash n})+\sum_{n'=1}^{N-1}\sum_{n=n'}^{N-1}I(Z_{n+1},Z_{n}|Z^{N-1}).
\end{align}
Finally, we rearrange this equation and divide both hand sides by $N$; we get the result.
\BlackBox

\section{Proof of Theorem~\ref{LM_bounds} in Linear Regression Model}
Here we show the results of Bayesian linear regression again. A Bayesian linear regression model is given as $p(Y|x,w)=\mathcal{N}(w^\top \phi(x),\beta^{-1})$ where $\phi(x):=(\phi_1(x),\dots,\phi_d(x))^\top\in\mathbb{R}^d$ is a $d$-dimensional feature vector of the input $x$ and each $\phi_i:\mathbb{R}^d\to\mathbb{R}$. We assume a prior distribution over $W$ as $P(W)=\mathcal{N}(0,\alpha^{-1} I_d)$. We define a design matrix as $\Phi=(\phi(x_1),\dots,\phi(x_N))^\top\in\mathbb{R}^{N\times d}$. We also define a target vector as $\mathbf{Y}=(Y_1,\dots,Y_N)^\top$. Then a posterior distribution is given by $P(W|Z^N)=\mathcal{N}(m_N,S_N)$ where $m_N=\beta S_N\Phi^\top \mathbf{Y}$ and $S_N^{-1}:=\alpha I_d+\beta \Phi^\top\Phi$. We also have a posterior predictive distribution as $P(Y|X,Z^N):=\mathcal{N}(m_N^\top\phi(x),\sigma^2_N(x))$ where $\sigma^2_N(x):=\beta^{-1}+\phi(X)^\top S_N\phi(X)$. Thus, the entropy of the posterior predictive distribution is given as
\begin{align}
H(Z|Z^N)=H(X)+H(Y|X,Z^N)=H(X)+\mathbb{E}_{p(X)}\frac{1}{2}\ln \sigma^2_N(X)+\frac{1}{2}d(\ln(2\pi)+1).
\end{align}
Thus, we can calculate $I_n$ as follows
\begin{align}
I(Z,Z_{n}|Z^{N\backslash n})= \mathbb{E}_{p(X)}\frac{1}{2}(\ln \sigma^2_{N\backslash n}(X)-\ln \sigma^2_N(X) ).
\end{align}
Note that since $\sigma^2_N(X)\leq \sigma^2_{N\backslash n}$ holds \citep{PRML}, $I(Z,Z_{n}|Z^{N\backslash n})$ is always positive. Then we use the relation
\begin{align}
\frac{x-y}{x+y}\leq \frac{1}{2}(\ln x-\ln y)\leq  \frac{1}{2}\frac{x-y}{y}.
\end{align}

Using these relations, we have
\begin{align}
    &\frac{1}{2}\frac{\sigma^2_{N\backslash n}(X)-\sigma^2_N(X)}{\alpha^{-1}\phi(x)^\top\phi(x)+\beta^{-1}}\leq\frac{\sigma^2_{N\backslash n}(X)-\sigma^2_N(X)}{\sigma^2_N(X)+\sigma^2_{N\backslash n}(X)}\leq  \frac{1}{2}(\ln \sigma^2_{N\backslash n}(X)-\ln \sigma^2_N(X))\leq  \frac{1}{2}\frac{\sigma^2_{N\backslash n}(X)-\sigma^2_N(X)}{\sigma^2_{N}(X)}.
\end{align}
Note that $S_N^{-1}:=\alpha I_d+\beta \Phi^\top\Phi$ and $\Phi^\top\Phi=\sum_n \phi(X_n)\phi(X_n)^\top$. Thus, we have $S_N^{-1}=S_{N-1}^{-1}+\beta\phi(X_N)\phi(X_N)^\top$.
From this relation, by using the Woodbury formula, we have
\begin{align}
S_{N-1}&=S_N+S_N\phi(\beta^{-1}- \phi(x_N)^\top S_N\phi(x_N))^{-1}\phi^\top S_N.
\end{align}
Then by definition, we have
\begin{align}
    \sigma^2_{N\backslash n}(X)-\sigma^2_N(X)=\phi(x)^\top (S_{N\backslash n}-S_N)\phi(x)).
\end{align}
Combining these results, we have
\begin{align}
    \sigma^2_{N\backslash n}(X)-\sigma^2_N(X)=\frac{(\phi(x)^\top S_N\phi(x_n))^2}{\beta^{-1}-\phi(x_n)^\top S_N\phi(x_n)}.
\end{align}
Note that $\beta^{-1}-\phi(x_n)^\top S_N\phi(x_n)$ is always positive. To confirm this, first, we regard this as the shur complement of the matrix
\begin{align}
M=\begin{pmatrix}
S_N^{-1} & \phi(x_n) \\
\phi(x_n)^\top & \beta^{-1} \\
\end{pmatrix}    
\end{align}
and use the fact that $\mathrm{det}(M)=\mathrm{det}(S_N^{-1})\mathrm{det}(\beta^{-1}-\phi(x_n)^\top S_N\phi(x_n))$. We can confirm that $M$ is positive definite because we can show that $v^\top M v >0 $, where $v\in\mathbb{R}^{N+1}$ is an arbitrary vector, using the fact that $S_N$ is positive definite. Then $\beta^{-1}-\phi(x_n)^\top S_N\phi(x_n)$ must be 
positive definite since $M$ and $S_N$ are positive definite.
Thus, we have
\begin{align}
    \frac{(\phi(x)^\top S_N\phi(x_n))^2}{\beta^{-1}-\phi(x_n)^\top S_N\phi(x_n)}\geq (\phi(x)^\top S_N\phi(x_n))^2.
\end{align}
In conclusion, we have
\begin{align}
    \frac{1}{N}\sum_{n=1}^N I(Z,Z_{n}|Z^{N\backslash n})\geq \mathbb{E}_{p(X)}\frac{1}{2N}\frac{1}{\alpha^{-1}\phi(x)^\top\phi(x)+\beta^{-1}}\sum_{n=1}^N\frac{ (\phi(x)^\top S_N\phi(x_n))^2}{\omega(x_n)},
\end{align}
where $\omega(x_n):=\beta^{-1}-\phi(x_n)^\top S_N\phi(x_n)$.

\section{Proof of the Asymptotic Result in Theorem~\ref{thm_asymp2}}\label{app_asymptotic}
We introduce the following definition,
\begin{align}
    \mathcal{G}_N(\alpha):=-\mathbb{E}_{W,Z^N,Z}\log\mathbb{E}_{\tilde{W}|Z^N}p(Z|W)^\alpha,
\end{align}
where $p(Z|W)^\alpha$ is the power of $\alpha\in\mathbb{R}$ of $p(Z|W)^\alpha$. We only consider the $\alpha$ such that $|\mathcal{G}_N(\alpha)|<\infty$. We also simplify the expression of the expectation since the distribution with which we take the expectation is clear. When $\alpha=1$, this is the Bayesian risk in the Bayesian learning setting. We also define
\begin{align}
    \mathcal{T}_N(\alpha):=-\mathbb{E}_{W,Z^N}\frac{1}{N}\sum_{n=1}^N\log\mathbb{E}_{\tilde{W}|Z^N}p(Z_n|W)^\alpha.
\end{align}
Compared to $\mathcal{G}_N(\alpha)$, this corresponds to the training error.

We then have the following relation
\begin{align}
    \mathcal{G}_{N-1}(\alpha=1)&=-\mathbb{E}_{W,Z^{N-1},Z_N}\log\mathbb{E}_{\tilde{W}|Z^{N-1}}p(Z_N|W)\nonumber\\
    &=\mathbb{E}_{W,Z^{N-1},Z_N}\left[\log\frac{1}{\mathbb{E}_W\prod_{n=1}^NP_{Z_n|W}}\mathbb{E}_W\left[ p(Z_N|W)^{-1}\prod_{n=1}^Np(Z_n|W)\right]\right]\nonumber\\
    &=\mathbb{E}_{W,Z^{N-1},Z_N}\log\mathbb{E}_{\tilde{W}|Z^N}p(Z_N|W)^{-1}\nonumber\\
    &=\mathbb{E}_{W,Z^{N-1},Z_N}\left[\frac{1}{N}\sum_{n=1}^N\log\mathbb{E}_{\tilde{W}|Z^N}p(Z_n|W)^{-1}\right]\nonumber\\
    &=-\mathcal{T}_N(\alpha=-1).
\end{align}

Recall that
\begin{align}
    I(W;Z|Z^N)=I(W;Z_N|Z^{N-1})-I(Z,Z_N|Z^{N-1}),
\end{align}
then we have
\begin{align}
    I(Z,Z_N|Z^{N-1})=\mathcal{G}_{N-1}(\alpha=1)-\mathcal{G}_{N}(\alpha=1),
\end{align}
thus we have
\begin{align}
    I(Z,Z_N|Z^{N-1})=-\mathcal{T}_N(\alpha=-1)-\mathcal{G}_{N}(\alpha=1).
\end{align}
Next we consider the Taylor expansion of $\mathcal{T}_N(\alpha)$ and $\mathcal{G}_{N}(\alpha)$ about $\alpha$. This expansion is studied asymptotically in \cite{watanabe2018mathematical} rigorously.
Using Theorem 3 in \cite{watanabe2018mathematical}, we have
\begin{align}
    I(Z,Z_N|Z^{N-1})=o\left(\frac{1}{N}\right),
\end{align}
where $o(\frac{1}{N})$ is little $o$.

Finally, from Theorem~\ref{thm_main_cmd}, we have
\begin{align}
    \frac{1}{N}\sum_{n'=1}^{N-1}\sum_{n=n'}^{N-1}I(Z_{n+1},Z_{n}|Z^{n-1})=-I(W;Y|X,Z^N)+\frac{1}{N}I(W;Z^N)-\frac{1}{N}\sum_n I_n\leq \frac{1}{N}I(W;Z^N).
\end{align}
Thus $\frac{1}{N}I(W;Z^N)=O(1/N)$, we obtain Eq.~\eqref{eq_order_info}.
\begin{remark}
The finite variance assumption is satisfied for regular models. We say that a statistical model is regular if its map $w\to p(z|w)$ is one-to-one, which implies $p(z|w_1)=p(z|w_2)\Rightarrow w_1=w_2$ and its Fisher information matrix is positive definite for arbitrary $w\in\mathcal{W}$. Intuitively, this means that when we have enough samples, the posterior distribution of a regular model converges to the Gaussian distribution. 
\end{remark}

\section{Proofs of Lemma~\ref{lem_cd_MI} and Theorem~\ref{eq_bayesian_generalization}}\label{Proof_cmi}
\proof
The conditional mutual information is rewritten as
\begin{align}
    &I(W;Y|X,Z^N)\nonumber \\
&=E_{W,Z^N,Z}\left[-\ln\mathbb{E}_{W|Z^N}p(Y|X,W)+\ln p(Y|X,W)]\right]\nonumber \\
&=E_{W,Z^N,Z}\left[-\ln\mathbb{E}_{W|Z^N}p(Y|X,W)\right]-H[Y|X,W].
\end{align}

On the other hand, the mutual information is rewritten as
\begin{align}
    &I(W;Z^N)\nonumber \\
    &=\mathbb{E}_{Z^N,W}[-\ln p(Z^N)+\ln p(Z^N|W)]\nonumber \\
    &=-\mathbb{E}_{Z^N}\ln p(Z^N)-H[Z^N|W]\nonumber \\
    &=-\mathbb{E}_{Z^N}\mathbb{E}_{W|Z^N}\ln p(Z^N|W)+\mathbb{E}_{Z^N}\mathrm{KL}(p(W|Z^N)|p(W))-H[Z^N|W].
\end{align}
This implies
\begin{align}
    -\mathbb{E}_{Z^N}\ln p(Z^N)=I(W;Z^N)+H[Z^N|W].
\end{align}
Note that $H[Z^N|W]=NH[Z|W]$ since the test and training data points are i.i.d. By combining these results, we get Lemma~\ref{lem_cd_MI}. Moreover, combining these relations with Theorem~\ref{thm_main_cmd}, we get Theorem~\ref{eq_bayesian_generalization}.
\BlackBox

\subsection{Relation to Bayesian Regret}\label{App_regret}
Here, we discuss how the Bayesian learning setting is related to Bayesian regret problems. Bayesian inference has been utilized for the sequential decision-making problem since we can incorporate the information from past observations into the posterior distribution. Here, we discuss how to utilize our developed theories for the sequential decision problem. We assume that, at each round $n$, we are given data $X_n$ and predict $Y_n$ using $Z^{n-1}$. When using an optimal decision, we suffer from loss 
 $R_l(Y_n|X_n,Z^{n-1})-R_l(Y_n|X_n,W)=\mathrm{MER}_l(Y_n|X_n,Z^{n-1})$ at  round $n$. We define an average cumulative regret as
\begin{align}
    \mathrm{Reg}_l(N):=\frac{1}{N}\sum_{n=1}^N\mathrm{MER}_l(Y_n|X_n,Z^{n-1}).
\end{align}
Under this definition, we obtain the following relation between regret and MER:
\begin{corollary}
\begin{align}\label{regret}
    &\mathrm{MER}_{\log}(Y|X,Z^N)=\mathrm{Reg}_{\log}(N)-\frac{1}{N}\sum_{n=1}^N I_n-\frac{1}{N}\sum_{n'=1}^{N-1}\sum_{n=n'}^{N-1}I_{n+1,n}.
\end{align}
\end{corollary}

\proof
Note that
\begin{align}
    -\ln p(Z_n|Z^{N-1})=-\ln \mathbb{E}_{W|Z^{N-1}}p(Z_n|W)=-\ln p(Z^{n})+\ln p(Z^{N-1}).
\end{align}
Thus, we have
\begin{align}
    \mathbb{E}_{Z^n}[-\ln p(Z_n|Z^{N-1})]=I(W;Z^n)-I(W;Z^{N-1})+H.
\end{align}
where $H$ is the entropy $H[Z|W]$. Here we omit the data point index from this entropy since all the entropy $H[Z_n|W]$ are equivalent since each $Z_n$ are i.i.d.  By adding from $n=1$ to $N$, we can reformulate a regret as
\begin{align}
    E_{Z^N}\sum_{n=1}^N[-\ln p(Z_n|Z^{N-1})]=I(W;Z^N)+NH.
\end{align}
This concludes the proof.
\BlackBox

\section{Proof of Theorem~\ref{inside_term}}
 We express the learned parameters as $\tilde{W}$, which follows the Bayesian posterior distribution $p(\tilde{W}|Z^N)$. Note that we have the Markov chain $W-Z^N-\tilde{W}$. Recall that $p(W|Z^N):=\frac{p(Z^N,W)}{p(Z^N)}$. Then we have
 \begin{align}
L(\tilde{W};Z^N|W)&=\mathbb{E}_{p(W)p(\tilde{Z}^N|W)}\mathbb{E}_{p(\tilde{W}|\tilde{Z}^N)p(Z^N|W)}\left[\log \frac{p(Z^N|W)p(\tilde{W}|W)}{p(Z^N,\tilde{W}|W)}\right]\nonumber \\
&=\mathbb{E}_{p(W)p(\tilde{Z}^N|W)}\mathbb{E}_{p(\tilde{W}|\tilde{Z}^N)p(Z^N|W)}\left[\log \frac{p(Z^N|W)}{p(Z^N|\tilde{W})}\right]\nonumber \\
&=NR_{\log}^{\mathrm{Gibbs}}-NH[Y|X,W].
\end{align}

\section{Proofs of Meta-learning in Theorem~\ref{thm_main_cmd_meta}}\label{app_meta}
The proof of the sensitivity in meta-learning is almost identical to that of Theorem~\ref{thm_main_cmd}. First, using Theorem~\ref{thm_main_cmd} we can decompose the MEMR as follows.
\begin{align}
    &I(Y;W|X,Z^N,Z^{NM})\nonumber \\
    &=\frac{I(W;Z^N|Z^{NM})}{N}-\frac{1}{N}\sum_{n=1}^N I(Z,Z_n|Z^{N\backslash n},Z^{NM})-\frac{1}{N}\sum_{n'=1}^{N-1}\sum_{n=n'}^{N-1}I(Z_{n+1},Z_{n}|Z^{n-1},Z^{NM}).
\end{align}
Here we applied Theorem~\ref{thm_main_cmd} to the meta-test test data and meta-test training dataset. Then we have
\begin{align}
    &I(Y;W|X,Z^N,Z^{NM})\nonumber \\
    &=\frac{I(W,U;Z^N|Z^{NM})}{N}-\frac{1}{N}\sum_{n=1}^N I(Z,Z_n|Z^{N\backslash n},Z^{NM})-\frac{1}{N}\sum_{n'=1}^{N-1}\sum_{n=n'}^{N-1}I(Z_{n+1},Z_{n}|Z^{n-1},Z^{NM})\nonumber \\
    &=\frac{I(W;Z^N|U)}{N}+\frac{I(U;Z^N|Z^{NM})}{N}-\frac{1}{N}\sum_{n=1}^N I(Z,Z_n|Z^{N\backslash n},Z^{NM})-\frac{1}{N}\sum_{n'=1}^{N-1}\sum_{n=n'}^{N-1}I(Z_{n+1},Z_{n}|Z^{n-1},Z^{NM}),
\end{align}
where we used the fact that $I(U;Z^N|Z^{NM})=0$, which results from the Markov chain $(U,Z^{NM})-W-Z^N$. Next, we use the following two lemmas, the modified versions of Lemma~\ref{lmm_entropuy} and \ref{MI_basic}, in the proof of Theorem~\ref{thm_main_cmd}.
\begin{lemma}\label{lmm_entropuy2}
For any $m\in (1,2,\dots, M]$, we have
\begin{align}
    H[Z^{N,(m+1)}|Z^{Nm}]=H[Z^{N,(m+1)}|Z^{N(m-1)}]-I(Z^{N,(m+1)};Z^{N,(m)}|Z^{N(m-1)}),
\end{align}
where $Z^{Nm}:=(Z^{N,(1)},\cdots,Z^{N,(m)})$ and $Z^{N(m-1)}:=(Z^{N,(1)},\cdots,Z^{N,(m-1)})$.
\end{lemma}
We can derive this by direct calculation. We also have the following lemma.
\begin{lemma}\label{MI_basic2}
For any $m\in (1,2,\dots, M]$
\begin{align}
    I(U;Z^{N,(m+1)}|Z^{Nm})=I(U;Z^{N,(m)}|Z^{N(m-1)})-I(Z^{N,(m+1)};Z^{N,(m)}|Z^{N(m-1)}).
\end{align}
\end{lemma}
We can prove this almost in the same way as Lemma~\ref{MI_basic}. Then, similarly to Eq.~\eqref{eq_derivation} in Theorem~\ref{thm_main_cmd}, we have
\begin{align}
    &I(U;Z^{NM})\nonumber \\
    &=\sum_{m=1}^MI(U;Z^{N,(m)}|Z^{N(m-1)})\nonumber \\
    &=MI(W;Z^{N,(M)}|Z^{N(M-1)})+\sum_{m'=1}^{M-1}\sum_{m=m'}^{M-1}I(Z^{N,(m+1)},Z^{N,(m)}|Z^{N(m-1)})\nonumber\\
    &=MI(W;Z^N|Z^{NM})+\sum_{m=1}^MI(Z^N,Z^{N,(m)}|Z^{N(M\backslash m)})+\sum_{m'=1}^{M-1}\sum_{m=m'}^{M-1}I(Z^{N,(m+1)},Z^{N,(m)}|Z^{N(m-1)}),
\end{align}
where $Z^{N(M\backslash m)}:=(Z^{N,(1)},\dots,Z^{N,(m-1)},Z^{N,(m+1)},\dots,Z^{N,(M)})$.
Combining these results, we have
\begin{align}
    &I(Y;W|X,Z^N,Z^{NM})\nonumber \\
    &=\frac{I(W;Z^N|U)}{N}+\frac{1}{NM}I(U;Z^{NM}) \\
    &-\frac{1}{NM}\sum_{m=1}^MI(Z^N,Z^{N,(m)}|Z^{N(M\backslash m)})-\frac{1}{NM}\sum_{m'=1}^{M-1}\sum_{m=m'}^{M-1}I(Z^{N,(m+1)},Z^{N,(m)}|Z^{N(m-1)})\\
    &-\frac{1}{N}\sum_{n=1}^N I(Z,Z_n|Z^{N\backslash n},Z^{NM})-\frac{1}{N}\sum_{n'=1}^{N-1}\sum_{n=n'}^{N-1}I(Z_{n+1},Z_{n}|Z^{n-1},Z^{NM}).
\end{align}

\section{Numerical Experiments}\label{app_experiments}
Here we show a detailed explanation of the numerical experiments in the main paper and additional results.
\subsection{Experiments in Bayesian Learning Setting}
Here we explain the settings in Sec.~\ref{sec_not_meta_exp}. We generated input $x~\sim \mathcal{N}(0, \mathbf{1}_{d'})$. We used $d'=1$ in the paper. We then set $\alpha=\beta=1$ in all the experiments. As the feature map, we used Gaussian feature map, defined as $\phi(x)=[e^{-\frac{1}{2}\|x-\mu_1\|^2},\cdots,e^{-\frac{1}{2}\|x-\mu_d\|^2}]$. We set $d=10$ and $\mu_1$ to $\mu_d$ evenly in the interval $[-2, 2]$.

\begin{figure}[tb!]
 \centering
 \includegraphics[width=0.5\linewidth]{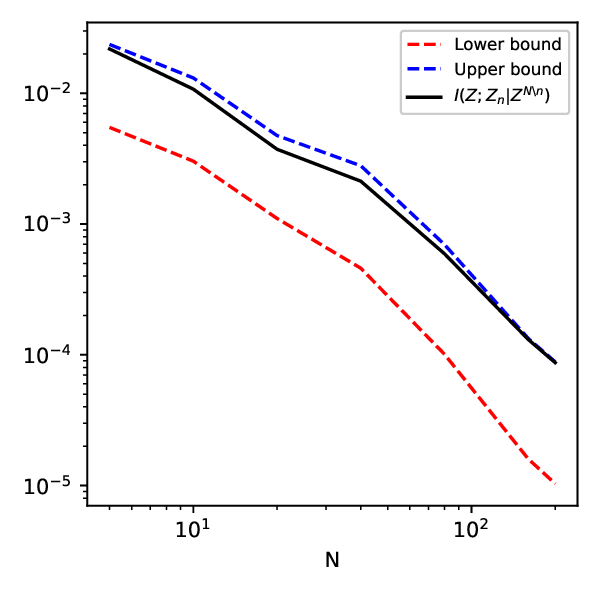}
 \vspace{-0.9truemm}
 \caption{The upper and lower bounds of the sensitivity derived in Theorem~\ref{LM_bounds}.}
\label{fig_ampm_bound}
\end{figure}

Here we show additional numerical results, the upper and lower bounds of $I_n$ derived in Theorem~\ref{LM_bounds}. The result is shown in Fig.~\ref{fig_ampm_bound}. We can see that the upper bound is very accurate, while the lower bound is relatively loose.

\begin{figure}[tb!]
 \centering
 \includegraphics[width=0.7\linewidth]{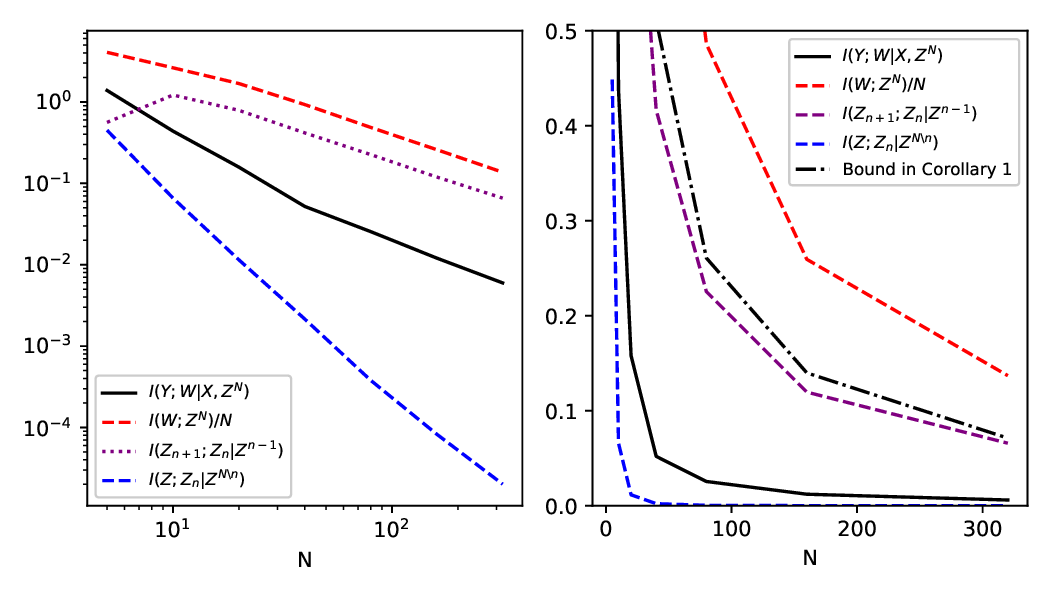}
 \vspace{-0.9truemm}
 \caption{The information-theoretic quantities appeared in Theorem~\ref{thm_main_cmd} and Corollary~\ref{col_improved_bound} for polynomial basis function.}
\label{fig_ampm_poly}
\end{figure}

Next, we show the results when using a polynomial feature map, $\phi(x)=[x,x^2,\cdots,x^5]$. We show the results in Fig.~\ref{fig_ampm_poly}. We can see that the information-theoretic quantities behave similarly to when we use the Gaussian basis function for the feature map.

\subsection{Experiments in Bayesian Meta-learning Setting}
Here we introduce the meta-learning settings of our numerical experiments.
For the prediction stage, we have the meta-test posterior predictive distribution as follows;
\begin{align}
p(Y;W|X,Z^N,Z^{NM})&=\int p(Y|X,W)p(W|Z^N,Z^{NM})dW \notag \\
&=\int p(Y|X,W)p(W|Z^N,U)p(U|Z^{NM})dWdU.
\end{align}
Note that from the meta-training dataset, we can learn the posterior of the hyper-prior $p(U)$ as $p(U|Z^{NM})$. We can learn the posterior of the parameter of the meta-task as $p(W|Z^N,U)$.

We are focusing on the linear regression model, and we put on a Gaussian prior on $W$ and Gaussian hyper-prior on $U$ as $p(W|U)=\mathcal{N}(U,\alpha^{-1}\mathbf{1})$ and $p(U)=\mathcal{N}(0,\gamma^{-1}\mathbf{1})$ where $\mathbf{1}$ is the identity matrix. Thus, we can analytically calculate all the related posterior and predictive distributions. Thus, we can analytically calculate the CMI and MI in the information-theoretic bound.

We first present the explicit form of $p(W|Z^N,U)$. Recall the definitions. A linear regression model is given as, $p(Y|X,W)=\mathcal{N}(W^\top \phi(x),\beta^{-1})$ where $\phi_t(x):=(\phi_1(x),\dots,\phi_d(x))^\top\in\mathbb{R}^d$ is a $d$-dimensional feature vector of the input $x$ of meta-test training dataset. We define a design matrix as $\Phi_t=(\phi(x_1),\dots,\phi(x_N))^\top\in\mathbb{R}^{N\times d}$. We also define a target vector as $\mathbf{y}=(y_1,\dots,y_N)^\top$. 
\begin{lemma} The posterior distribution of the meta-task parameter is given as Gaussian distribution, $p(W|Z^N,U)=\mathcal{N}(m_t,S_t)$ where
\begin{align}
&m_t:=S_t(\beta \Phi_t^\top \mathbf{y}+\alpha U),\\
&S_t^{-1}:=\alpha I_d+\beta \Phi_t^\top\Phi_t.
\end{align}
\end{lemma}
\proof
The  joint distribution $p(U,Z^N,W)$ is Gaussian; thus, $p(W|Z^N,U)$ is Gaussian distribution. So we only need to calculate $\log p(U,Z^N,W)$ and focus on the quadratic and linear terms of $W$ since those coefficients characterize the mean and variance of the Gaussian distribution. Then we have
\begin{align}
-\log p(U,Z^N,W)=W^\top (\alpha I_d+\beta \Phi^\top\Phi)W +2W^\top(\beta\Phi^\top \mathbf{y}+ \alpha U)+\mathrm{Const}.
\end{align}
By re-arranging this, we obtain the result.
\BlackBox

Next, we present the posterior distribution of hyperparameter $U$. We introduce the definition of the design matrix of the $m$-th meta-training dataset $\phi_m(x):=(\phi_1(x),\dots,\phi_d(x))^\top\in\mathbb{R}^d$ and a design matrix as $\Phi_m=(\phi_m(x_1),\dots,\phi_m(x_N))^\top\in\mathbb{R}^{N\times d}$. We use the same feature map for the meta-training and meta-testing stages. Under this setting, we have the following result.
\begin{lemma} The posterior distribution of hyper-parameter is given as the Gaussian distribution, $p(U|Z^{NM})=\mathcal{N}(m_u,S_u)$ where
\begin{align}
&m_u:=S_u\sum_{m=1}^M \mathbf{y}_m^\top s_m\Phi_m, \\
&S_u^{-1}:=\sum_{m=1}^M (\gamma\mathbf{1}+\Phi_m^\top s_m\Phi_m),\\
&s_m^{-1}:=\beta^{-1}\mathbf{1}+\alpha^{-1}\Phi_m\Phi_m^\top.
\end{align}
\end{lemma}
\proof
We consider the joint distribution for the meta-training as $p(U,Z^NM,W^M)=p(Z^{NM}|W^M)p(W^M|U)p(U)$. First, we can easily integrate out $W$ from the joint distribution. Then, for the $m$-th task, $\int p(Z^{N,m}|W_m)p(W_m|U=u)dW_m$ is given as the Gaussian distribution $\mathcal{N}(\Phi_m u,\beta^{-1}\mathbf{1}+\alpha^{-1}\Phi_m\Phi_m^\top)$. 
Since the joint distribution $p(U,Z^{NM})$ is the Gaussian, $p(U|Z^{NM})$ is also the Gaussian distribution. Thus, we only need to calculate $\log p(U,Z^{NM})$ and focus on the quadratic and linear terms of $U$. Then we have
\begin{align}
-\log p(U,Z^{NM})=U^\top (\gamma+\sum_{m=1}^M \Phi_m^\top s_m\Phi_m)U +2U^\top(\sum_{m=1}^M \Phi^\top s_m\mathbf{y}_m)+\mathrm{Const},
\end{align}
where $s_m^{-1}:=\beta^{-1}\mathbf{1}+\alpha^{-1}\Phi_m\Phi_m^\top$ and $\mathbf{y}_m$ is the target variables of the $m$-th task.
Thus, we obtain the result.
\BlackBox

Finally, we calculate the meta-test predictive distribution. This is also given as the Gaussian distribution.
\begin{lemma} The posterior predictive distribution of the meta-test is given as the Gaussian distribution, $p(Y;W|X,Z^N,Z^{NM})=\mathcal{N}(m_f,S_f)$ where
\begin{align}
&m_f=\beta\mathbf{y}^\top \Phi_tS_t\phi(x)+\alpha m_u^\top S_t\phi(x),\\
&S_f=\beta^{-1}+\phi(x)^\top S_t\phi(x))+(\alpha S_t \phi(x))^\top S_u (\alpha S_t \phi(x)).
\end{align}
\end{lemma}
\proof
We first calculate $\int p(Y|X,W)p(W|Z^N,U)p(U|Z^{NM})dW$. This is given as the Gaussian distribution $\mathcal{N}(m_t^\top\phi(x),\beta^{-1}+\phi(x)^\top S_t\phi(x))$, where we defined $m_t$ and $S_t$ in the above. Then applying the expectation formula of the Gaussian distribution, we have
\begin{align}
&m_f=\mathbb{E}_U[m_t^\top\phi(x)]=\mathbb{E}_U[(S_t(\beta \Phi_t^\top \mathbf{y}+\alpha U))^\top \phi(x)]=\beta\mathbf{y}^\top \Phi_tS_t\phi(x)+\alpha m_u^\top S_t\phi(x),\\
&S_f=\beta^{-1}+\phi(x)^\top S_t\phi(x))+(\alpha S_t \phi(x))^\top S_u (\alpha S_t \phi(x)).
\end{align}
\BlackBox

As for the experimental settings, we set $\gamma=1$. All the other settings are the same as the not meta-learning settings.

\begin{figure}[tb!]
 \centering
 \includegraphics[width=0.8\linewidth]{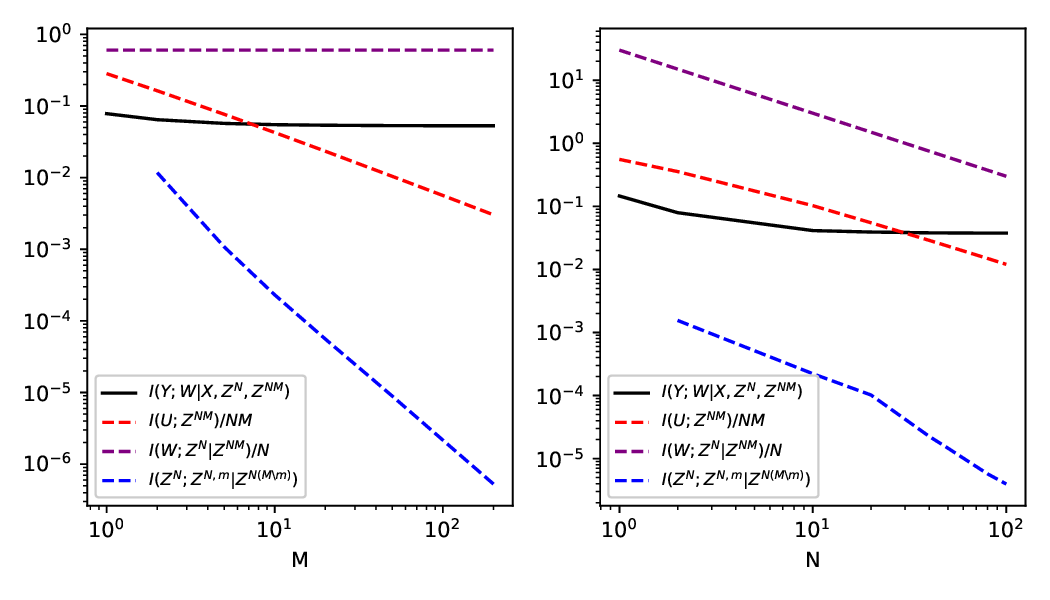}
 \vspace{-0.9truemm}
 \caption{The information-theoretic quantities appeared in Theorem~\ref{thm_main_cmd_meta} for polynomial basis function.}
\label{fig_ampm_poly_meta}
\end{figure}

In the main paper, we showed the results of the Gaussian feature map. Here we show additional experimental results about the polynomial feature map. In Fig.~\ref{fig_ampm_poly_meta}, we show the results when the polynomial feature map, $\phi(x)=[x,x^2,\cdots,x^5]$ is used. We can see that the information-theoretic quantities behave similarly to when we use the Gaussian basis function for the feature map.

\end{document}